\documentclass[10pt,twocolumn,letterpaper]{article}

\usepackage{cvpr}              % To produce the CAMERA-READY version
\usepackage[accsupp]{axessibility}  % Improves PDF readability for those with disabilities.

\usepackage{url}
\usepackage{graphicx}
\usepackage{xspace}
\usepackage{multirow}
\usepackage{graphicx}
\usepackage{subcaption}
\usepackage{soul}  % for highlighting text
\usepackage{makecell}

\newcommand{\methodname}{CoReDiT\xspace}

\definecolor{cvprblue}{rgb}{0.21,0.49,0.74}
\usepackage[pagebackref,breaklinks,colorlinks,allcolors=cvprblue]{hyperref}
\usepackage[capitalize]{cleveref}

\def\paperID{9} % *** Enter the Paper ID here
\def\confName{CVPR}
\def\confYear{2026}

\title{\methodname: Spatial Coherence-Guided Token Pruning and Reconstruction for Efficient Diffusion Transformers \vspace{-5pt}}

\author{Zhuojin Li\thanks{Work done during internship at Qualcomm AI Research.} \quad Hsin-Pai Cheng \quad Hong Cai \quad Shizhong Han \quad Fatih Porikli\\
Qualcomm AI Research\thanks{Qualcomm AI Research is an initiative of Qualcomm Technologies, Inc.}\\[-2pt]
{\tt\small \{zhuoli,hsinpaic,hongcai,shizhan,fporikli\}@qti.qualcomm.com}\\[-2pt]
}

\begin{document}
\maketitle

\begin{abstract}

Diffusion Transformers (DiTs) deliver remarkable image and video generation quality but incur high computational cost, limiting scalability and on-device deployment.
We introduce \methodname, a structured token pruning framework for DiTs across vision tasks.
\methodname uses a linear-time \emph{spatial coherence} score to estimate local redundancy in the latent token lattice and skips high-coherence (redundant) tokens in self-attention.
To maintain a dense representation and avoid visual discontinuities, we reconstruct skipped attention outputs via coherence-guided aggregation of spatially neighboring retained tokens.
% that have participated in self-attention computation.
% 
We further introduce a progressive, block-adaptive pruning schedule that increases pruning gradually and allocates larger budgets to blocks and denoising steps with higher redundancy.
Across state-of-the-art diffusion backbones including PixArt-$\alpha$ and MagicDrive-V2, \methodname achieves up to \textbf{55\%} self-attention FLOPs reduction and inference speedups of \textbf{1.33$\times$} on cloud GPUs and \textbf{1.72$\times$} on mobile NPUs, while maintaining high visual quality.
Notably, \methodname also increases on-device memory headroom, enabling higher-resolution generation.
%
% Our results demonstrate that spatial coherence is a powerful signal for structured pruning in diffusion transformers.

\end{abstract}

\vspace{-5pt}
\section{Introduction}
% \vspace{-5pt}

% Background: Success of diffusion transformer
Diffusion models learn to synthesize data through an iterative denoising process,
and have achieved state-of-the-art results across various generative applications, such as conditional image generation, image editing (inpainting and super-resolution) and video synthesis.
Building on this success, Diffusion Transformers (DiTs)~\cite{peebles2023scalable} replace convolutional U-Nets with attention-based backbones that effectively model long-range dependencies among tokenized patches; this architecture scale effectively with model size and facilitate the incorporation of conditioning from various modalities (e.g., class labels, text, and depth maps).

% Key Challenges: Computational cost
However, these benefits come with demanding computational cost.
Self-attention scales quadratically with the number of tokens,
% which poses substantial challenges as image resolutions or number of video frames increase, especially on mobile platforms with tight memory budgets and constrained computational capacity.
making high-resolution images and multi-frame videos particularly expensive and often impractical on compute- and memory-constrained mobile platforms.
%
% Opportunities: redundant tokens
Empirically, a large fraction of tokens correspond to visually redundant or low-saliency regions (e.g., uniform backgrounds and smooth textures), suggesting substantial room for compute reduction.
Prior work reduces inference cost by pruning or merging tokens using attention-based saliency~\cite{kim2022learned,singh2024tosa}, feature similarity~\cite{bolya2022token}, or learned routing/predictors~\cite{rao2021dynamicvit,you2025layer}.
%
% i.e., selecting only the most informative tokens to participate in attention,
%
% such as saliency signals (e.g., attention scores~\cite{kim2022learned,singh2024tosa}), similarity (e.g., ToME~\cite{bolya2022token}) and learnable predictors (e.g., DynamicViT~\cite{rao2021dynamicvit}, DiffCR~\cite{you2025layer}).
%
% The key insight is that not all tokens contribute equally to attention: typically only a small set of tokens carry the semantic, such as texture, edge, while a large portion of patches encode redundant background or low-saliency content. 
%
% By saving computations on the uninformative tokens, existing methods can substantially reduce FLOPs and memory while preserving visual semantics.
%
However, several challenges limit their applicability to diffusion inference:
(1) localizing low-saliency tokens efficiently and effectively,
(2) preserving visual semantics for the tokens that do not participate in attention,
and (3) determining a pruning schedule that adapts across transformer blocks and denoising timesteps.

\noindent \textbf{Contributions:} To address these challenges, we propose \methodname, a token pruning framework for accelerating \emph{pre-trained} DiTs that targets self-attention while preserving output fidelity.
\methodname consists of three components:
(1) \textit{Spatial coherence-based selector} (\cref{section:token_selection}):
We observe that redundant tokens in low-saliency regions are highly similar to their spatial neighbors.
To exploit this, we partition the token lattice into small, non-overlapping grids and estimate a \emph{spatial coherence} score for each token, based on its feature similarity to tokens in the same grid.
This score can be implemented with linear-time, hardware-friendly reductions (e.g., grid means and dot products), adding minimal overhead while reliably selecting redundant tokens (with high coherence) to bypass attention.
%
% A key intuition is that the tokens in low-saliency regions are similar to the spatially neighboring tokens;
% %
% we divide each entire image into multiple grids and define a spatial coherence score that effectively depicts the similarity of a token to other tokens in the same grid. The tokens with highest spatial coherence score will bypass the attention computation.
%
%
(2) \textit{Coherence-based reconstruction} (\cref{section:token_reconstruction}):
Skipping tokens can break local spatial continuity, leading to artifacts in the generated outputs. To preserve a dense token lattice for subsequent blocks, we reconstruct skipped tokens from retained neighbors using similarity-weighted aggregation within a local neighborhood.
This content-aware interpolation (unlike zeroing or naive forwarding) preserves visual semantics and mitigates artifacts while maintaining compatibility with the original DiT architecture.
(3) \textit{Progressive, block-adaptive pruning} (\cref{section:progressive_pruning}):
Pruning tolerance varies across blocks and timesteps: different blocks carry varying semantic importance, and late diffusion steps are typically less tolerant than early ones~\cite{you2025layer}.
Hence, we fine-tune with a progressive schedule that \emph{automatically allocates} the pruning budget to the most redundant blocks.
At regular intervals during fine-tuning, we estimate a timestep-weighted redundancy score for each transformer block and increment the pruning ratio of the block with the highest redundancy;
%
% This concentrates pruning on blocks where the model is more resilient,
%
% this progressive schedule introduces capacity reduction gradually to ensure stability.
%
gradually increasing the pruning ratio enables fast recovery after each update and ensures stable adaptation.

Combining these components results in an efficient pruning pipeline that reduces attention computation while preserving high-fidelity generation.
Across text-to-image and video generation,
% (including the autonomous driving model MagicDrive-V2)
\methodname enables higher-resolution synthesis and stable conditional alignment, achieving up to 55\% reduction in self-attention FLOPs and wall-clock speedups of 1.33$\times$ on cloud GPUs and 1.72$\times$ on mobile devices.

% Contributions
% Below summarizes our contributions:
% \begin{itemize}
% \item We propose and validate spatial coherence as a pre-attention redundancy signal, showing it correlates with token saliency across DiT/ViT backbones and diffusion timesteps while adding negligible overhead.

% \item Instead of simply forwarding, we provide a coherence-based reconstruction that preserves locality and texture for skipped tokens, which effectively preserves the visual semantics and avoids artifacts.

% \item We design a progressive, block-adaptive schedule that concentrates pruning blocks with high redundancy and protects late diffusion steps, improving stability versus uniform ratios.

% \item Combing these components yields a simple pipeline that reduces attention computation while maintaining high-fidelity visual semantics.
% %
% Comprehensively evaluations demonstrate that our approach is hardware-friendly, and can achieve consistent saves computational cost across DiT architectures and tasks.

% \end{itemize}
\section{Related Work}
% \vspace{-5pt}
% Key distinguish from existing work:
% 1) We perform fine-tuning on pre-trained models (instead of training from scratch)
% 2) We target on hardware-efficient algrotihm

\textbf{Diffusion Models.} Diffusion models have demonstrated impressive generative performance across image, video, and 3D domains.
%
% , and controllable editing, delivering strong fidelity, compositionality, and robustness under diverse conditioning signals (text, layout, pose, depth) [CITE].
%
Early works commonly adopt U-Net backbones~\cite{ronneberger2015u,ho2020denoising},
%, which couples multi-scale convolutional encoders/decoders with skip connections to propagate fine detail while denoising across noise levels.
%
% This design proved effective for 2D imagery and moderate resolutions, yet it inherits inductive biases and memory traffic patterns that can limit scalability and global context modeling.
%
% To reduce computational cost and enable high-resolution synthesis, diffusion models often operate in a compressed latent space: an autoencoder maps pixels to a low-dimensional latent grid where denoising occurs, then decodes back to pixel space~\cite{rombach2022high}.
%
and latent diffusion~\cite{rombach2022high} improves scalability by denoising in a compressed latent space.
%
% Latent diffusion preserves perceptual quality while cutting FLOPs and memory, and it cleanly integrates cross-attention for text or other conditional signals [CITE].
%
% More recently, diffusion transformers (DiTs)~\cite{peebles2023scalable} have emerged as alternatives to U-Net. These models patchify latents and use attention to capture long-range dependencies with a uniform block structure, demonstrating strong scalability for higher resolutions, longer videos, and multi-modal conditioning.
%
Recently, Diffusion Transformers (DiTs)~\cite{peebles2023scalable} replace U-Net architectures with stacks of transformer blocks, offering better scalability to high resolution and flexible conditioning, but incurring high inference cost from quadratic attention and multi-step denoising.

\textbf{Efficient Diffusion Transformers.}
% Due to the quadratic cost in memory and computation, a substantial body of work has focused on designing efficient DiT models:
Prior studies have improved the efficiency of DiTs along several complementary directions:
(1) \textit{Token reduction within a timestep}.
% Prior Works reduce the effective token either by merging redundant latents in a training-free way (e.g., ToMe variants~\cite{bolya2023token}) or by learning importance scores/keep-ratios that vary by layer and timestep~\cite{you2025layer}.
Training-free token reduction often merges or downsamples tokens to reduce the effective sequence length, including token merging methods (e.g., ToMe~\cite{bolya2023token}) and diffusion-specific variants (e.g., ToMA~\cite{lu2025toma}) that aim to reduce merge overhead.
Other training-free strategies reduce attention cost by downsampling only key/value tokens (e.g., ToDo~\cite{smith2024todo}).
Learning-based methods learn per-layer/timestep keep ratios or token routers for content generation, including DiffCR~\cite{you2025layer} and dynamic token-density schemes such as FlexDiT~\cite{chang2024flexdit}.
A key challenge in DiTs (vs.\ ViTs) is that naively discarding tokens can introduce spatial discontinuities (texture breaks, boundary artifacts), motivating approaches that either preserve structure via merge/unmerge or explicitly reconstruct skipped content.
(2) \textit{Cross-timestep caching and feature reuse}.
%
% Feature-reuse methods~\cite{wimbauer2024cache} exploit the smooth evolution of hidden states across denoising steps, reusing block/layer activations with policies that decide when to refresh.
%
Caching-based methods exploit temporal redundancy across denoising steps by reusing intermediate features with selective refresh~\cite{wimbauer2024cache,lv2024fastercache}.
Recent DiT-specific works further cache at finer granularity (e.g., token-wise feature caching in ToCa~\cite{zou2024accelerating}).
%
%
% Later work~\cite{lv2024fastercache} brings this to DiTs with token-wise or layer-wise selection and adds forecasting/correction to mitigate drift.
%
% Beyond per-token reuse, cluster-driven approaches cache/propagate representative tokens within spatial clusters (e.g., ClusCa~\cite{zheng2025clusca}), and hybrid schemes combine clustering, pruning, and caching policies (e.g., CAT Pruning~\cite{cheng2025catpruning}).
%
These approaches primarily target \emph{cross-step redundancy} and are conceptually orthogonal to within-step token reduction.
\textit{(3) Dynamic computation and conditional routing.}
Dynamic DiT architectures adjust compute across timesteps and/or spatial regions, e.g., DyDiT~\cite{zhao2024dynamic} uses timestep-wise dynamic width and spatial token skipping, while DiffCR~\cite{you2025layer} fine-tunes token routers with differentiable compression ratios.
Compared to token reduction, these methods introduce explicit routing modules and typically optimize a dynamic compute graph.
\textit{(4) Sparse/structured attention kernels.}
Another line accelerates DiTs by replacing full attention with sparse attention patterns and hardware-efficient kernels,
including training-free profiling-based sparse attention (e.g., Sparse VideoGen~\cite{xi2025sparse}) and trainable sparse attention (e.g., VSA~\cite{zhang2025vsa}).
These methods change the attention operator itself rather than reducing the token set, and can be complementary to token pruning/merging.
(5) \textit{Weight-pruning and architecture editing}.
Beyond tokens, structured and unstructured pruning remove channels, heads, or even full DiT blocks, with brief recovery fine-tuning or lightweight calibration to retain quality~\cite{fang2023structural}.
Other work (e.g., grafting~\cite{chandrasegaran2025exploring}) edits the architectures of pretrained DiTs to explore more efficient backbones under small compute budgets; these approaches are naturally complementary to token sparsity methods.

\textbf{Positioning of \methodname.}
\methodname is a \emph{post-training} method that targets \emph{within-timestep} redundancy in DiT latent grids: it performs \emph{structured token skipping} guided by local spatial coherence, and \emph{reconstructs} skipped tokens to preserve a full lattice for subsequent layers.
This differs from caching-based methods (e.g., FasterCache/ToCa) that exploit \emph{cross-timestep reuse}, from dynamic routing approaches (e.g., FlexDiT/DiffCR) that introduce learnable routing policies, and from token-merging methods (e.g., ToMe) that explicitly merge/unmerge tokens.
Compared to purely training-free schemes, \methodname uses lightweight fine-tuning to learn block/timestep-specific pruning schedules to improve hardware-relevant speed/quality trade-offs.

% \textbf{Token Pruning.} Token pruning is an inference-time acceleration strategy that skips computation on less important tokens (patch embeddings) in a transformer.
% %
% Existing token pruning work can be categorized into saliency signals (e.g., attention scores~\cite{kim2022learned,singh2024tosa}), similarity (e.g., ToME~\cite{bolya2022token}) and learnable predictors (e.g., DynamicViT~\cite{rao2021dynamicvit}).
% %
% Existing token pruning methods have achieved promising results in ViTs for simple tasks (e.g., classification and object detection)~\cite{liang2022not}, where the objective primarily focus on key tokens and tolerate token dropping.
% %
% In DiT, by contrast, naively skipping tokens can result in visual discontinuities in the generated images (e.g., inconsistent textures, distorted boundaries).
% %
% Recent DiT-specific work dynamically modulates token density across layers/timesteps (e.g., FlexDiT~\cite{chang2024flexdit}, DiffCR~\cite{you2025layer}).

\section{Proposed Approach: \methodname}\label{section:methods}
% \vspace{-5pt}
% \subsection{Overview}\label{section:method_overview}
% This can just be an overview paragraph without having a subsection title. I will comment out the subsection titles of motivation in the following as well with a similar rationale :) -hc

\begin{figure}[t]
    \centering
    \includegraphics[width=\linewidth]{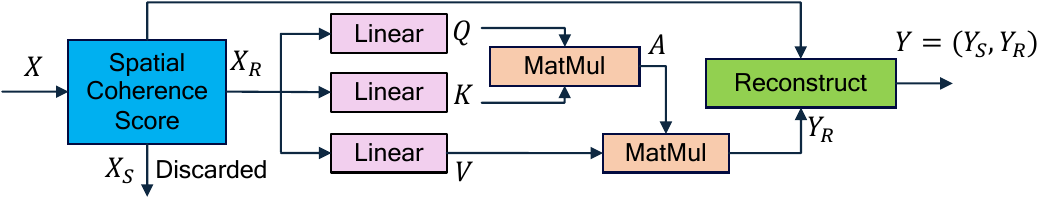}
    % \vspace{-5pt}
    \caption{\small Workflow of \methodname. In each transformer block, input tokens $X$ are partitioned into retained tokens $X_R$ and skipped tokens $X_S$ using the proposed spatial coherence score that captures \emph{local redundancy} (\cref{section:token_selection}).
    Only the retained tokens $X_R$ participate in self-attention, reducing attention compute by skipping $X_S$.
    We then reconstruct the missing attention outputs $Y_S$ from nearby retained outputs $Y_R$ via coherence-guided aggregation (\cref{section:token_reconstruction}), preserving a dense token lattice for subsequent blocks.
    Finally, we learn an adaptive pruning-ratio schedule via coherence-guided progressive pruning during lightweight fine-tuning (\cref{section:progressive_pruning}).
    }\label{fig:pipeline_overview}
\end{figure}

\cref{fig:pipeline_overview} summarizes the workflow of \methodname:
In each transformer block, we partition the input tokens (patch embeddings) $X$ into a retained set $X_R$ and a skipped set $X_S$ according to a spatial coherence score that quantifies local redundancy of each token; only $X_R$ participates in multi-head self-attention, producing $Y_R = \text{MHSA}(X_R)$ (\cref{section:token_selection}).
%
% For skipped tokens $X_S$, we synthesize their attention outputs from nearby retained tokens via coherence-guided aggregation $Y_S(i) = \sum_{j \in \mathcal{N}(i; X_R)} w_{i, j} Y_R(j)$, where $\mathcal{N}(i; X_R)$ denotes retained tokens from neighboring spatial positions of skipped token $x_i$, and weights $w_{i, j} \propto \text{sim}(x_i, x_j)$ represents the similarity of token pair.
%
For skipped tokens $X_S$, we synthesize their attention outputs $Y_S$ using nearby retained outputs $Y_R$ with coherence-guided weights (\cref{section:token_reconstruction}).
Importantly, \emph{attention among retained tokens remains global}; locality is used only for efficient selection and reconstruction, not to impose a local-attention pattern.
To determine which block to prune more aggressively, we learn a progressive, block-adaptive pruning schedule during fine-tuning of a pretrained DiT, based on the estimated redundancy statistics (\cref{section:progressive_pruning}).

% \vspace{-2pt}
\subsection{Token Selection}\label{section:token_selection}
% \vspace{-5pt}

\begin{figure*}[t]
    \centering
    \begin{subfigure}[t]{.49\linewidth}
        \centering
        \includegraphics[width=.75\linewidth]{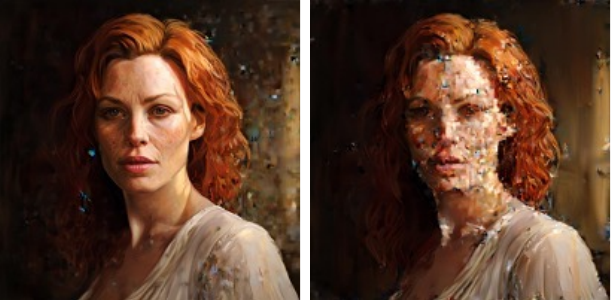}
        \caption{\small Token pruning with attention scores-based importance: removing tokens with the lowest scores (left) vs. the highest scores (right).}\label{fig:motivation_attn_score}
    \end{subfigure}
     \hfill
     \hfill
    \begin{subfigure}[t]{.49\linewidth}
        \centering
        \includegraphics[width=.92\linewidth]{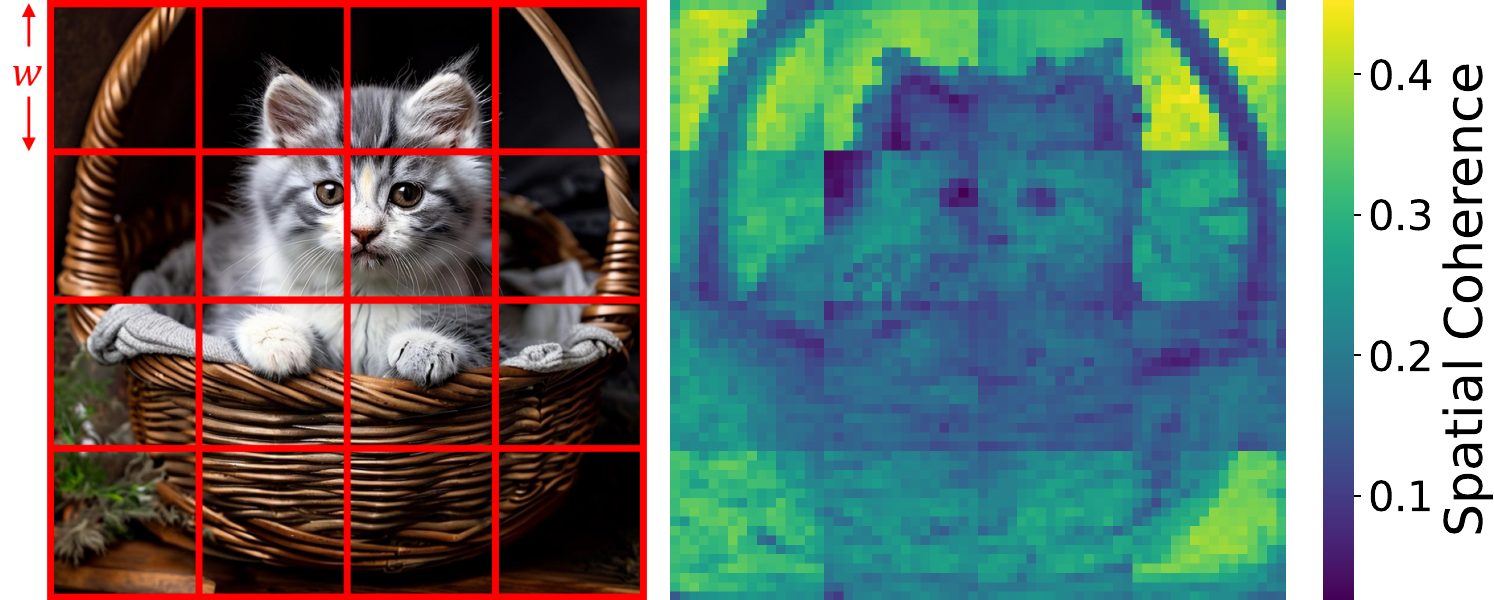}
        \caption{\small Example generated from PixArt-$\alpha$-1024. Left: image with grid partition overlay. Right: computed SC scores over $64\times 64$ tokens.}
        \label{fig:spatial_coherence}
    \end{subfigure}
    % \vspace{-5pt}
    \caption{\small Token selection motivation and visualization.}\label{fig:pruning_highest_lowest_attn_score}
    % \vspace{-5pt}
\end{figure*}

% \begin{figure}
%     \centering
%     \includegraphics[width=1.0\linewidth]{figs/patch_to_score.pdf}
%     \caption{spatial coherence score}
%     \label{fig:spatial_coherence}
% \end{figure}

\paragraph{Motivation.}
% \vspace{-5pt}

A common token pruning criterion is attention score-based importance~\cite{kim2022learned,singh2024tosa}:
Let $A = \text{softmax}(\frac{Q K^T}{\sqrt{d_k}})$ denote the attention matrix, where the column-sum in $A$ measures the total attention a token receives.
%
% Accordingly, we conduct experiments of pruning tokens with highest v.s. lowest attention scores for a DiT, as shown in \cref{fig:motivation_attn_score}.
%
% As can be seen, pruning tokens with high attention score severely impacts the visual semantics, since these tokens concentrate on salient image regions (e.g., edges, textures, object boundaries); in contrast, pruning tokens with low attention score results in smaller quality degradation because they primarily cover low-salience background areas and contribute less to the transformation of salient tokens.
%
As \cref{fig:motivation_attn_score} shows, pruning high-attention tokens severely damages visual semantics (salient regions), whereas pruning low-attention tokens tends to be less harmful (background regions).
However, computing global attention score $A$ requires quadratic time and memory in token count, which substantially offsets pruning gains.
Our key observation is that low-saliency tokens typically exhibit \emph{strong local redundancy}: their embeddings are highly similar to spatially nearby tokens.
Motivated by this, we design an efficient selection score based on local similarity that avoids forming the full $N \times N$ attention matrix.
% between a token and its neighboring tokens within each local region, which, as we shall see, is both effective and incurs minimal computation overhead.

% \vspace{-2pt}
\paragraph{Spatial Coherence (SC).}
% \vspace{-5pt}
Given $N$ tokens $X = \{x_i\}_{i=1}^N$ reshaped into a 2D spatial lattice $H \times W$ (given $N=HW$), we partition tokens into non-overlapping spatial grids of size $w \times w$ to estimate local (rather than global) similarity (\cref{fig:spatial_coherence}).
For a token $x_i$, define its spatial coherence (SC) as the mean cosine similarity to tokens in the same grid:
\begin{equation}
\text{SC}(x_i) := \frac{1}{|\mathcal{G}(i)|}\sum_{j \in \mathcal{G}(i)} \text{sim}(x_i, x_j)
\end{equation}
where $\mathcal{G}(i)$ indexes tokens in the same grid as token $x_i$. Specifically, we adopt cosine similarity $\text{sim}(x_i, x_j) = \frac{x_i^T x_j}{||x_i||_2 ||x_j||_2}$ due to its computational efficiency and strong empirical performance in prior work~\cite{bolya2022token}.
Averaging within each grid normalizes SC values across blocks with different $w$,
enabling comparisons in our block-adaptive schedule (\cref{section:progressive_pruning}).
%
% Note that the reason why partition tokens into grids is to capture the local spatial coherence among the image patches.
% -- please revise or move the above as needed. We should explain why we are doing the partition. -hc

% Efficiency improvement
\paragraph{Efficient computation and overhead.}
Naively computing within-grid pairwise similarities costs $O(w^2 N)$ dot products.
Instead, let $\hat{x_i} = \frac{x_i}{||x_i||_2}$ denote the normalized token embedding, and $\hat{g_i} = \frac{1}{|\mathcal{G}(i)|} \sum_{j \in \mathcal{G}(i)} \hat{x_j} $ denote the mean normalized embeddings within the grid of token $x_i$.
Then we can compute the spatial coherence score efficiently:
\begin{equation}
\text{SC}(x_i) = \frac{1}{|\mathcal{G}(i)|}\sum_{j \in \mathcal{G}(i)} \hat{x_i}^T \hat{x_j} = \frac{\hat{x_i}^T}{|\mathcal{G}(i)|} \sum_{j \in \mathcal{G}(i)} \hat{x_j} = \hat{x_i}^T \hat{g_i}.
\end{equation}
Essentially, the spatial coherence is an inner product between a token's normalized embedding and the mean normalized embedding of its grid.
%
% This reduces the complexity of score computation to $O(N)$, regardless of grid size $w$, since $g_i$ can be computed once per grid and reused for all tokens in that grid.
%
Thus, SC can be computed with one reduction per grid plus one dot product per token, i.e., $O(N)$ time, which is linear in token count, regardless of grid size $w$.

\paragraph{Selection rule.}
High SC indicates a token is well-explained by nearby patches (redundant), so we skip the top-$K$ tokens by SC:
\begin{equation}
X_S = \text{Top}_K \left(\{\text{SC}(x_i)\}_{i=1}^N\right), \quad X_R = X\setminus X_S
\end{equation}
where $K=r(l,s)\cdot N$ is determined by the pruning ratio $r(l,s)$ for block $l$ and timestep $s$.
Empirically, as shown in \cref{fig:spatial_coherence}, high-SC tokens (brighter colors) concentrate in backgrounds, while boundaries/textures tend to have lower SC.
Note that our proposed spatial coherence is not only used for token selection, but also utilized to update skipped tokens and guide our progressive pruning (\cref{section:progressive_pruning}).

% Visualization
% Figure~\cref{fig:spatial_coherence} visualizes the spatial coherence score on a pretrained PixArt-$\alpha$-1024 model.
%
% For $64 \times 64$ input tokens, we choose grid size 16 or 9 to partition the tokens into $4 \times 4$ or $7 \times 7$ grids.
%
% As can be seen, high-coherence scores concentrate in low-saliency regions that are locally redundant, while edges and textured structures exhibit low coherence.
%
% Accordingly, our selection mechanism skips tokens with highest spatial coherence score in a transformer block to preserve visual quality: $X_S = \{x \in X\ |\ \text{token}\ x\ \text{has the top-$K$ spatial coherence values} \}$. Note that our proposed spatial coherence is not only used for token selection, but also utilized to update skipped tokens and guide our progressive pruning, as we discuss in the following.

\subsection{Token Reconstruction}\label{section:token_reconstruction}
% \vspace{-5pt}
\paragraph{Motivation.}

% \begin{figure}
%     \centering
%     \includegraphics[width=.25\linewidth]{figs/src/token_skipping_artifact.pdf}
%     \caption{Visual artifacts when simply skipping tokens}
%     \label{fig:token_skipping_artifact}
% \end{figure}

\begin{figure}[t]
    \centering
    \begin{subfigure}[t]{.32\linewidth}    
        \centering
        \includegraphics[width=.95\linewidth]{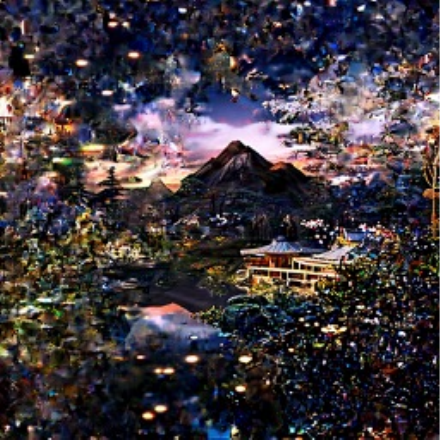}
        \caption{\small Simply skipping tokens}
        \label{fig:token_skipping_artifact}
    \end{subfigure}
    \begin{subfigure}[t]{.32\linewidth}
        \centering
        \includegraphics[width=.95\linewidth]{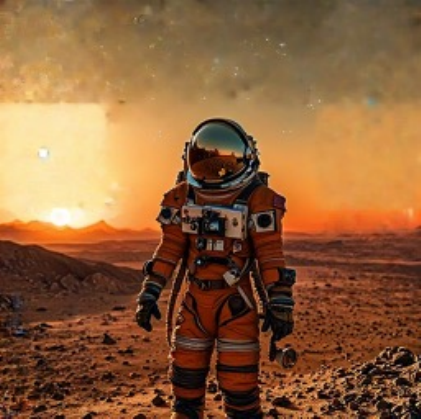}
        \caption{\small Visual artifacts on border}\label{fig:border_visual_artifact}
    \end{subfigure}
    % \hspace{0.25em}
    \begin{subfigure}[t]{.32\linewidth}
        \centering
        \includegraphics[width=1.28\linewidth]{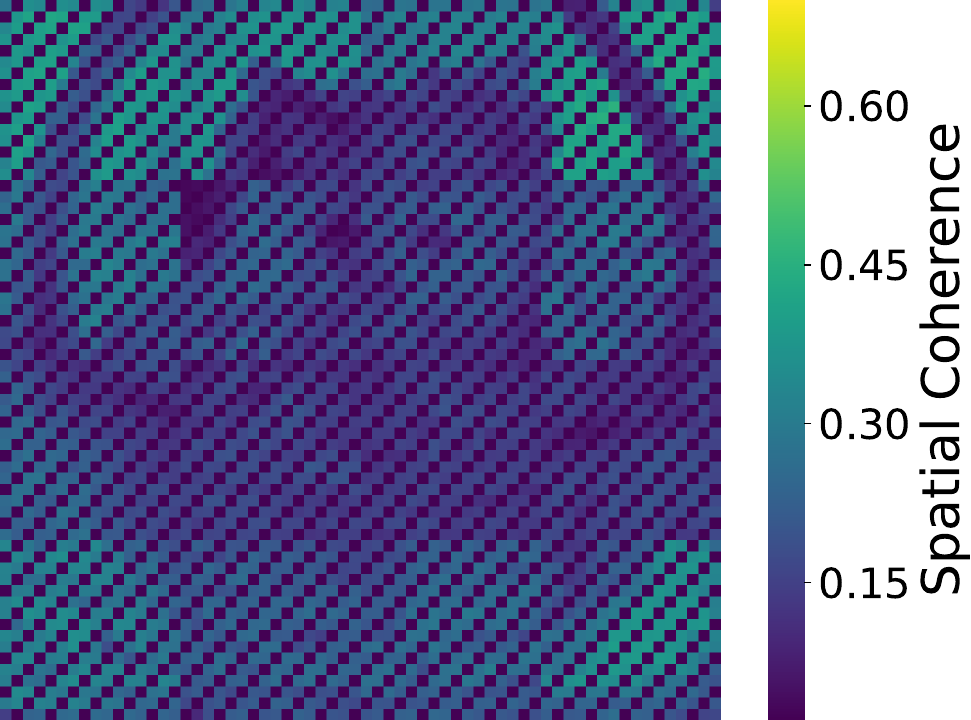}
        \caption{\small $m$-stride token retention}\label{fig:protected_tokens}
    \end{subfigure}
    % \vspace{-5pt}
    \caption{\small Issues and proposed improvements. (a) Naively skipping tokens causes inconsistent textures, which can be mitigated by token reconstruction. (b) Using a fixed grid size during partitioning causes visual artifacts along grid borders; alternating grid sizes across consecutive blocks mitigates these artifacts. (c) $m$-stride token retention ensures at least one token is retained in each sub-grid to serve as a reconstruction anchor.}\label{fig:improments}
    % \vspace{-10pt}
\end{figure}

% Issue of simply skipping
% Token pruning has achieved promising results in ViTs for simple tasks such as classification and object detection~\cite{liang2022not}, where the objectives primarily focus on key tokens and can tolerate token dropping.
%
% In DiT, in contrast, naively skipping tokens can result in visual discontinuities in the generated images (e.g., inconsistent textures, distorted boundaries), as shown in~\cref{fig:token_skipping_artifact}.
%
Unlike ViTs for discriminative tasks (e.g., classification and object detection) where dropping tokens can be tolerated~\cite{liang2022not}, generative DiTs require spatially coherent representations at every block and timestep.
Naively skipping tokens in MHSA can introduce discontinuities (inconsistent textures, distorted boundaries), as in \cref{fig:token_skipping_artifact}.
%
% Consequently, to preserve visual semantics of the generated images, we propose to reconstruct the transformed version of the skipped tokens based on the retained tokens that have gone through self-attention in the local neighborhood, by leveraging their correlation relationship.
% 
% In particular, since the proposed selection mechanism is inclined to skip redundant tokens with high coherence, these tokens can be more effectively reconstructed.
%
We therefore reconstruct the missing attention outputs for skipped tokens using nearby retained outputs, preserving a dense lattice for subsequent blocks.
Because our selector preferentially skips \emph{redundant} (high-SC) tokens, these tokens are particularly well-suited to reconstruction from neighbors.

\paragraph{Coherence-based reconstruction.}
% \vspace{-5pt}
% To compute the block transformation $Y$ from input tokens $X$, we identify a skipped set $X_S$ and forward only the retained tokens $X_R = X - X_S$ through multi-head self-attention to obtain $Y_R = \text{MHSA}(X_R)$.
%
% To preserve spatial continuity without performing attention on $X_S$, we synthesize the missing transformations $Y_S$ by similarity-weighted aggregation over nearby retained tokens: $\forall x_i \in X_S$,
%
Let $X_R$ be tokens retained for attention and $Y_R=\text{MHSA}(X_R)$. For a skipped token $x_i\in X_S$, a natural reconstruction is similarity-weighted aggregation over nearby retained tokens:
%
% \begin{equation}
% % \forall x_i \in X_S,\quad
% y_i = \frac{\sum_{j \in \mathcal{S}(i)} \text{sim}(x_i, x_j) \cdot y_j}{\sum_{k \in \mathcal{S}(i)}\text{sim}(x_i, x_k)} \approx \frac{\sum_{j \in \mathcal{S}(i)} \text{SC}(x_j) \cdot y_j}{\sum_{k \in \mathcal{S}(i)}\text{SC}(x_k)}
% \end{equation}
\begin{equation}
y_i
= \frac{\sum_{j \in \mathcal{S}(i)} \text{sim}(x_i, x_j)\, y_j}{\sum_{k \in \mathcal{S}(i)} \text{sim}(x_i, x_k)}
\label{eq:reconstruct_exact}
\end{equation}
%
% where $\mathcal{S}(i) = \{j \ | \ j \in \mathcal{G}(i) \land x_j \in X_R\}$ is the set of nearby retained token indexes.
%
where $\mathcal{S}(i)$ denotes retained tokens in a local neighborhood of $x_i$.
Intuitively, the transformation result of a token should be more alike to that of a token with a higher similarity.
However, computing $\text{sim}(x_i,x_j)$ per pair adds non-trivial overhead.
%
% This process involves $O(w^2)$ pairwise similarities per token, resulting in overall complexity of $O(w^2 N)$.

\paragraph{Approximate $\text{sim}(x_i,x_j)$ with $\text{SC}(x_j)$.}
Skipped tokens are selected to have \emph{high} coherence within their grid (\cref{section:token_selection}), meaning $\hat{x_i}$ is close to the grid mean $\hat{g_i}$.
Formally, define $\epsilon_i = \|\hat{x_i}-\hat{g_i}\|_2$; high-SC implies small $\epsilon_i$. Then for any $j\in\mathcal{G}(i)$:
\begin{equation}
\text{sim}(x_i,x_j)=\hat{x_i}^T\hat{x_j}
=\hat{g_i}^T\hat{x_j} + (\hat{x_i}-\hat{g_i})^T\hat{x_j}
=\text{SC}(x_j) + O(\epsilon_i)
\label{eq:sc_proxy}
\end{equation}
since $\|\hat{x_j}\|_2=1$ bounds the residual term by $\epsilon_i$.
Thus, for the tokens we prune (high-SC, small $\epsilon_i$), using $\text{SC}(x_j)$ yields a close approximation to similarity-based weights while avoiding pairwise similarity computation.
We therefore approximate \cref{eq:reconstruct_exact} as:
\begin{equation}
y_i \approx
\frac{\sum_{j \in \mathcal{S}(i)} \text{SC}(x_j)\, y_j}{\sum_{k \in \mathcal{S}(i)} \text{SC}(x_k)}.
\label{eq:reconstruct_sc}
\end{equation}

\paragraph{Sub-grids and complexity.}
If $\mathcal{S}(i)$ spans an entire grid, \cref{eq:reconstruct_sc} may over-smooth and reduce spatial variation.
We address this by subdividing each $w\times w$ grid into smaller \emph{sub-grids} of size $w_s\times w_s$ and defining $\mathcal{S}(i)$ as retained tokens in the \emph{same sub-grid} as $x_i$.
This preserves locality at boundaries while keeping computation linear.
Reconstruction uses only local weighted sums over at most $w_s^2$ tokens per skipped token; with small $w_s$ (e.g., 3), the overhead is modest and scales linearly with token count, i.e., $O(N)$.
Notably, this locality is \emph{only} for reconstruction; MHSA among $X_R$ is still global, so \methodname does not impose a local attention pattern.
%
% To improve the efficiency, we replace $\text{sim}(x_i, x_j)$ by $\text{SC}(x_j)$ to make reconstructions based more on the retained tokens with higher spatial coherence. 
% %
% However, this can result in the same transformation result of a skipped token within each grid, because of the weighted average of the transformation results of retained tokens.
% %
% To relieve this issue, we restrict aggregation within a smaller areas by dividing each grid into smaller sub-grids with size $w_s \times w_s$ (e.g., $4 \times 4$ or $3 \times 3$) to leverage the spatial locality of the image, which effectively reduces the complexity to $O(N)$.
% %
% %\begin{equation}
% %\forall x_i \in X_S,\quad y_i = \frac{\sum_{j \in \mathcal{S}(i)} \text{SC}(x_j) %\cdot y_j}{\sum_{k \in \mathcal{S}(i)}\text{SC}(x_k)}
% %\end{equation}

\paragraph{Micro designs.}
% \vspace{-5pt}
% Issue: visual artifacts
(1) Alternating grid size:
Fixed grid boundaries can introduce border artifacts due to limited inter-grid mixing in reconstruction (\cref{fig:border_visual_artifact}).
We mitigate this by alternating grid sizes across blocks (e.g., for $64\times 64$ tokens, alternate grid size $16\times 16$ and $9\times 9$), so boundaries shift and information can propagate across grid borders over consecutive blocks.
%
% Additionally, we observe visual artifacts at grid border (as shown in~\cref{fig:border_visual_artifact}) due to lack of inter-grid information exchange. We mitigate this issue by alternating grid sizes across transformer blocks.
%
% Specifically, for $32 \times 32$ tokens, we alternate grid size of $16 \time 16$ and $9 \time 9$.
%
This strategy enables cross-border information flow over consecutive blocks, eliminating visible discontinuities.
%
% Issue: skipp entire sub-grid
(2) $m$-stride token retention:
In highly redundant regions, an entire sub-grid could be skipped, leaving no anchors for reconstruction.
We therefore enforce retaining at least one token per $m$ positions ($m\le w_s$) by protecting a deterministic stride pattern.
For block index $l$, we retain tokens at spatial coordinates $[i,j]$ satisfying $(i+j-l)\bmod m = 0$ by subtracting a large offset from their SC so they are never selected into $X_S$.
%
% In some scenarios (e.g., large chunk of background), all tokens in a sub-grid might be skipped, which leads to no retained tokens as reconstruction reference.
% %
% To address the this issue, we enforce always retaining a token for every $m$ tokens ($m \le w_s$).
% %
% Specifically, for transformer block with index $l$, we enforce the tokens at position $[i, j]$ with $(i+j-l) \mod m = 0$ to be retained, by subtracting a large offset on their spatial coherence scores, e.g., $m=3$ in~\cref{fig:protected_tokens}.

\subsection{Pruning Ratio Schedule}\label{section:progressive_pruning}

\paragraph{Motivation.}

\begin{figure}[t]
    \centering
    \includegraphics[width=\linewidth]{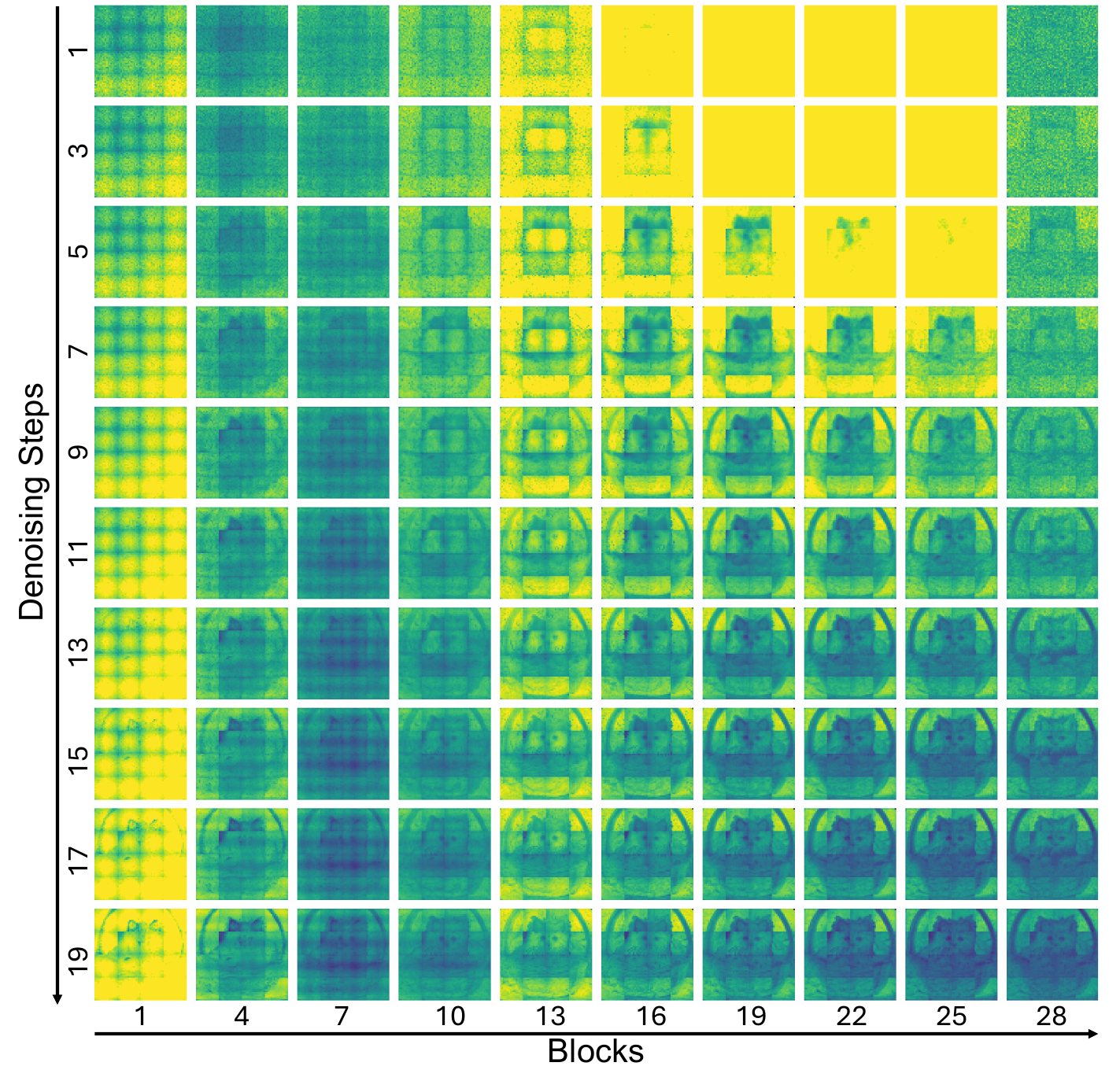}
    \caption{\small SC score evolution across transformer blocks and denoising timesteps (brighter = higher SC) for generating the image in~\cref{fig:spatial_coherence}. (a) Early timestep in the first block: representations are noise-dominated and exhibit high local redundancy. (b) Early timestep in the last block: weaker spatial structure and reduced locality. (c) Late timestep in the first block: refined spatial details with clearer object-level coherence. (d) Late timestep in the last block: partial refinement with moderate spatial structure. These patterns indicate that redundancy varies with both timestep and block, motivating adaptive pruning ratios across depth and denoising phase.}
    \label{fig:full_spatial_coherence}
\end{figure}

% Block-wise and timestep-wise
% The diffusion process maps noise to an image through sequential denoising steps.
%
Token redundancy varies across both denoising timesteps and transformer blocks.
%
% As shown in \cref{fig:coherence_score_across_block_and_timestep},
% %
\cref{fig:full_spatial_coherence} depicts the evolution of spatial coherence across denoising steps and blocks.
%
% As shown, early denoising steps are noise-dominated and establish global semantics, while late steps refine fine details such as textures and boundaries.
%
As shown, early steps display broadly elevated coherence (coarse structure), mid steps accentuate background redundancy, and late steps concentrate low-coherence pockets around edges and details~\cite{chang2024flexdit,you2025layer};
besides, within a timestep, different blocks exhibit different degrees of locality~\cite{chandrasegaran2025exploring}.
%
% These observations suggest that a uniform pruning ratio across timesteps and blocks is suboptimal.
%
% This progression motivates our block-adaptive schedule: we allocate larger pruning budgets where coherence remains high and throttle pruning where coherence drops, preventing over-pruning of emerging details and preserving final image fidelity.
%
% In addition, we empirically observe that applying a large pruning ratio abruptly can be hard to recover for DiT.
% We therefore adopt progressive pruning during fine-tuning, gradually increasing pruning budgets while prioritizing blocks/timesteps with higher measured redundancy~\cite{yang2023global}.
%
% Given that we prune tokens from pre-trained DiTs,
Motivated by this, we propose adapting the pruning ratio across blocks based on the statistics observed from real data.
Specifically, we adopt progressive pruning during fine-tuning: gradually increasing pruning budgets while prioritizing blocks/timesteps with higher measured redundancy.
This progressive design also conforms to the empirical finding in~\cite{yang2023global}: Applying a large pruning ratio abruptly to a pretrained model can be hard to recover, while progressive pruning leads to better results at comparable pruning levels.

\paragraph{Coherence-based progressive pruning.}
% \vspace{-5pt}
% 1) Block-wise decision
Let $K(l, s) = r(l, s) \cdot N$ denote the number of tokens pruned in block $l$ at denoising step $s$.
During fine-tuning, we increase $K(l, s)$ in increments of $\Delta k$ every $T$ training iterations until reaching a target budget.
To decide which block to increase next, we sort the SC scores in block $l$ in descending order, $\{\text{SC}_i^l\}_{i=1}^{N}$.
% , and define
% $R_l(K) = \sum_{i = 1}^K \text{SC}_i^l$.
%
We estimate the redundancy of the \emph{next} $\Delta k$ candidate tokens as:
\begin{equation}
% \Delta R_l = R_l(K_{l, s} + \Delta k) - R_l (K_{l, s})  
\Delta R_l \;=\; \sum_{i=K(l,s)+1}^{K(l,s)+\Delta k}\text{SC}_i^l,
\label{equ:initial_score}
\end{equation}
where $\Delta R_l$ is computed on the current minibatch.
%
% Note: the following sentence is moved to emprical study
% For example, \cref{fig:initial_score} presents the initial redundancy score for pre-trained PixArt-$\alpha$-1024.
%
Every $T$ fine-tuning iterations, we assign the next increment $\Delta k$ to the block with largest $\Delta R_l$, i.e., the block whose next pruned tokens are most redundant.

% 2) Timestep-wise decision
\paragraph{Timestep-wise decay.}
Pruning schedule should also reflect denoising phase: Early timesteps are noise-dominated and set global structure (more redundancy), while late timesteps refine local details (less redundancy).
Accordingly, we adopt more aggressive pruning early and reduce pruning late.
In principle, one could learn a distinct pruning ratio for each denoising step.
However, changing $r(l,s)$ across steps changes how many tokens are retained and therefore modifies the execution path of MHSA and reconstruction.
Implementing a fully step-specific schedule would require instantiating (and often caching) many step-dependent computation graphs and associated buffers (e.g., on mobile NPUs), which increases memory footprint and engineering complexity.
We therefore adopt a lightweight two-phase decay that captures the major redundancy shift from early (noise-dominated) to late (detail-refinement) denoising, inspired by prior studies~\cite{chang2024flexdit,you2025layer}, while requiring only \emph{two} graphs per block and keeping memory overhead low:
%
% Since each distinct per-timestep ratio would instantiate a different computational graph (which is costly for mobile platforms), we empirically adopt a two-phase simple schedule over denoising step $s$:
%
$
r(l, s) = 
\begin{cases}
    r(l), & 1 \le s \le s_0 \\
    c \cdot r(l), & s_0 < s \le S
\end{cases}
$
where $r(l)$ is the base pruning ratio for block $l$, $S$ is the total number of denoising steps, $s_0$ is the phase boundary,
% (e.g., $s_0 =15$ when $S = 20$)
and $0 < c < 1$ is the late-phase decay.
%
% Aligned with prior empirical observations~\cite{chang2024flexdit,you2025layer}, this timestep-wise decay $c < 1$ captures phase-dependent redundancy while requiring only two computational graphs per block.
%
% This schedule captures phase-dependent redundancy with only two computational graphs for each denoising step, reducing the overhead of compute and memory cost on mobile devices.

\begin{figure}[t]
    \centering
    \begin{subfigure}[t]{\linewidth}
        \centering
        \includegraphics[width=.85\linewidth]{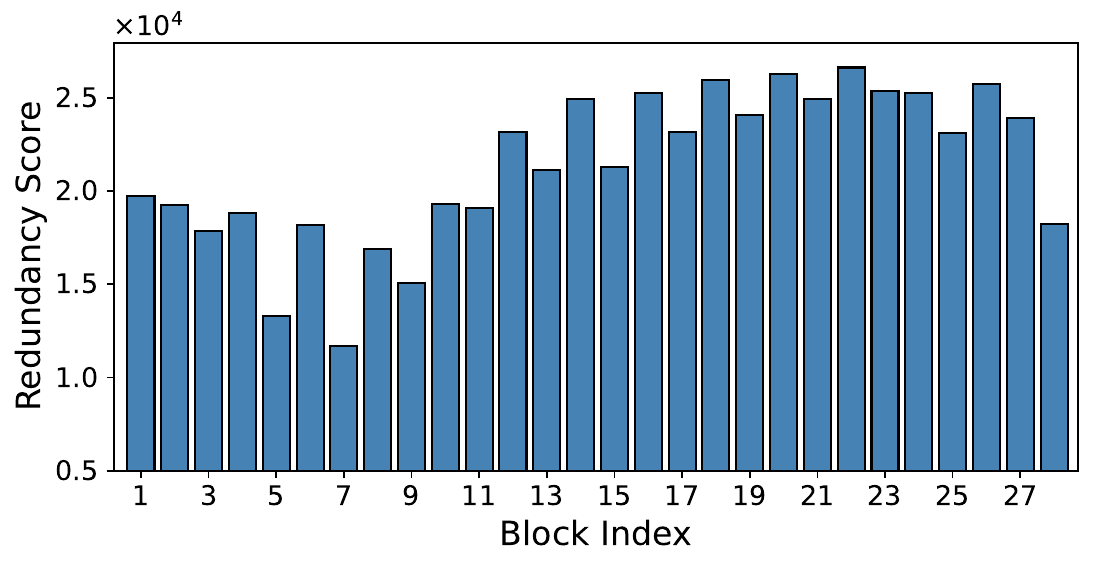}
        % \vspace{-16pt}
        \caption{\small 
        % Initial redundancy scores for the next $\Delta k$ tokens based on \cref{equ:initial_score} across transformer blocks.
        Block-wise redundancy of the next $\Delta k$ candidate pruned tokens at initialization (computed via \cref{equ:initial_score}).
        }\label{fig:initial_score}
    \end{subfigure}
    \hfill
    \begin{subfigure}[t]{\linewidth}
        \centering
        \includegraphics[width=.85\linewidth]{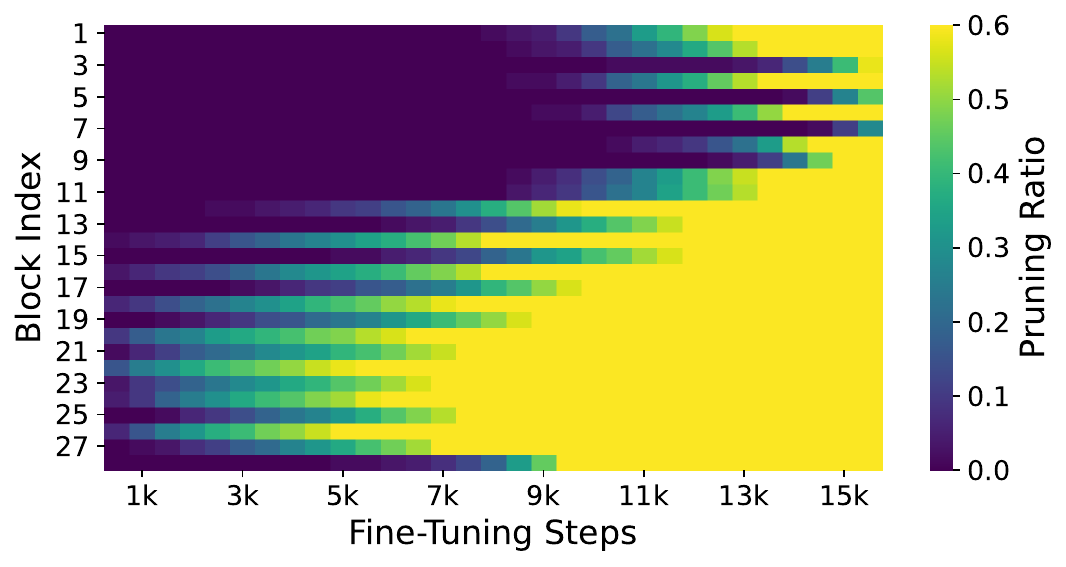}
        % \vspace{-5pt}
        \caption{\small
        % Evolution of pruning ratios over fine-tuning, updated every T steps based on block-wise redundancy.
        Learned pruning ratios over fine-tuning.
        }\label{fig:progressive_pruning_ratio}
    \end{subfigure}
    % \vspace{-5pt}
    \caption{\small Coherence-guided progressive pruning across transformer blocks for PixArt-$\alpha$-1024. Higher redundancy blocks receive larger pruning ratios as fine-tuning proceeds.}\label{fig:progressive_pruning_process}
%\vspace{-10pt}
\end{figure}

\begin{figure}[t]
    \centering
    \begin{subfigure}[t]{\linewidth}
        \centering
        \includegraphics[width=.85\linewidth]{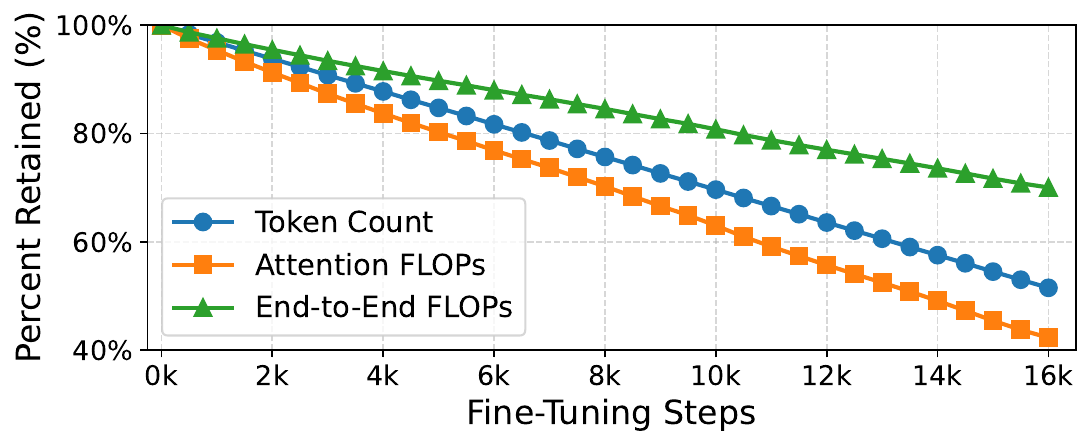}
        % \vspace{-5pt}
        \caption{\small Efficiency during progressive pruning: retained-token ratio, attention FLOPs, and end-to-end FLOPs.}\label{fig:progressive_pruning_flops}
    \end{subfigure}
    \newline
    \vspace{1em}
    % \hfill
    \begin{subfigure}[t]{\linewidth}
        \centering
        \includegraphics[width=.85\linewidth]{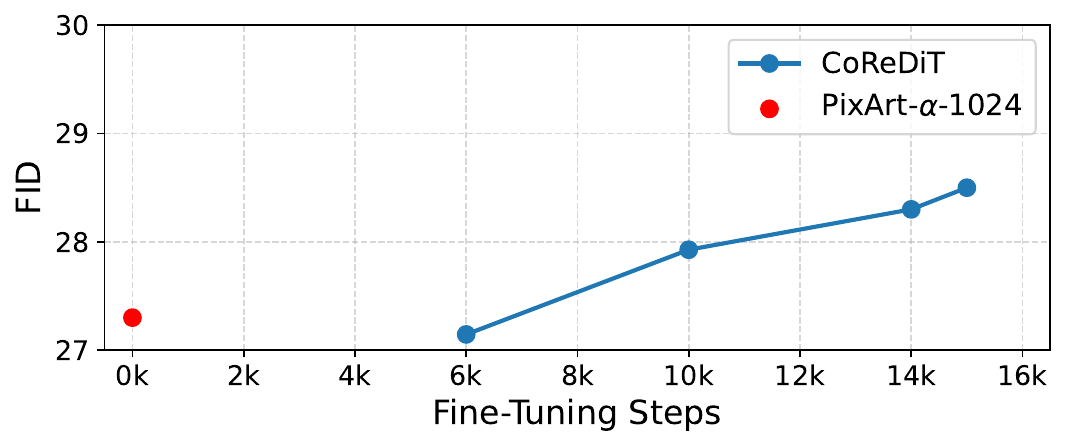}
        % \vspace{-5pt}
        \caption{\small Quality during progressive pruning (FID). Red dot: pretrained baseline; blue curve: \methodname during fine-tuning.
        % The red dot represents the original model, while the blue line corresponds to the proposed \methodname method.
        }\label{fig:progressive_pruning_fid}
    \end{subfigure}
    
    \caption{\small Progressive pruning dynamics for PixArt-$\alpha$-1024: efficiency gains and quality trend over fine-tuning.}
    \label{fig:progressive_pruning_metrics}
%\vspace{-10pt} 
\end{figure}

% Analysis on progressive pruning
\paragraph{Empirical study.}
We demonstrate progressive pruning on PixArt-$\alpha$-1024 by increasing the pruned tokens by $\Delta k = 64$ every $T = 15$ fine-tuning steps, with a per-block cap of 60\% tokens.
\cref{fig:initial_score} shows that blocks differ substantially in redundancy at initialization. Consequently, the learned schedule (\cref{fig:progressive_pruning_ratio}) allocates higher pruning ratios to consistently redundant blocks while keeping low-redundancy blocks lightly pruned (e.g., blocks 5 and 7).
% indicates, the schedule adapts to these differences, directing more pruning to blocks with higher redundancy.
%
% For example, blocks 5 and 7 exhibit low redundancy scores and therefore maintain low pruning ratios throughout the fine-tuning.
%
As a result, \methodname rapidly reduces attention and end-to-end FLOPs (\cref{fig:progressive_pruning_flops}) while maintaining generation quality (\cref{fig:progressive_pruning_fid}).

\section{Results}
% \vspace{-5pt}
\subsection{Experimental Setup}
% \vspace{-5pt}
% 1) Training Dataset
\paragraph{Models and training datasets.}
%
% 1-1) text-to-image
For text-to-image generation, we apply \methodname to the official checkpoints of PixArt-$\alpha$-1024~\cite{chen2023pixart} and PixArt-$\Sigma$-2048~\cite{chen2024pixart}.
We perform lightweight fine-tuning with progressive pruning on the 10K synthetic dataset from CLEAR~\cite{liu2024clear}, where images are generated by FLUX.1-dev~\cite{flux2024}.
%
% 1-2) video generation
For video generation, we apply \methodname to MagicDrive-V2~\cite{gao2024magicdrive} and fine-tune the released third-stage checkpoint on nuScenes~\cite{caesar2020nuscenes}.

% 2) Training configs
\paragraph{Progressive pruning configuration.}
% 2-1) text-to-image
For PixArt-$\alpha$-1024, we fine-tune with batch size of 20 on a single Nvidia H100 GPU; forPixArt-$\Sigma$-2048, we use a total batch size of 40 across 8 Nvidia H100 GPUs.
Unless otherwise stated, 
we update the pruning schedule every $T=15$ fine-tuning steps by pruning additional $\Delta k = 64$ tokens in the selected block, with a per-block pruning ratio cap of $60\%$;
we use a two-phase timestep schedule with decay $c=0.25$ and $S - s_0 = 5$, i.e., decay only in the final 5 denoising steps, as detailed in \cref{section:progressive_pruning}.
Following CLEAR~\cite{liu2024clear}, we apply distillation during fine-tuning using both model output loss $L_{\text{pred}}$  and the pruned-block loss $L_{\text{attn}}$:
$
L_{distill} = L_{ori} + \alpha L_{pred} + \beta L_{attn}
$
, following their hyperparameters $\alpha = \beta = 0.5$.
%
% 2-2) video generation
For MagicDrive-V2, we prune only the \emph{spatial} transformer blocks, since they represent the main computational bottleneck, with a significantly higher token count than the temporal blocks.
We fine-tune the third-stage checkpoint with SP size of 8 across 8 Nvidia H100 GPUs.
% , with learning rate of $2.5 \times 10^{-6}$.
%
% For 30 denoising steps $s$, we apply a pruning ratio $r(l, s)$ such that $r(l, s) = r(l)$ for the first 26 steps and $r(l, s) = 0.25 \cdot r(l)$ for the final 4 steps.

% 3) Evaluation
% 3-1) text-to-image
\paragraph{Evaluation metrics.} For text-to-image generation, following CLEAR~\cite{liu2024clear}, we evaluate on 10k caption-image pairs randomly sampled from MSCOCO 2014 validation~\cite{lin2014microsoft} using FID~\cite{heusel2017gans}, CLIP text similarity~\cite{radford2021learning}, and Inception Score (IS)~\cite{salimans2016improved}.
%
% 3-2) video generation
For video generation, we report FVD~\cite{unterthiner2018towards}, LPIPS~\cite{zhang2018unreasonable}, PSNR, and SSIM~\cite{wang2004image} for video quality; we also evaluate conditional alignment using BEVFormer~\cite{li2022bevformer}  with 
mAP and mIoU.

\subsection{Main Results: PixArt-$\alpha$-1024}
% \vspace{-5pt}

\begin{table*}[t]
\centering
\small
\begin{tabular}{|c|c|c|c|c|c|}
\hline
\multirow{2}{*}{\textbf{Model}} & \multicolumn{2}{c|}{\textbf{FLOPs Reduction}} & \multicolumn{3}{c|}{\textbf{Image Quality}} \\ \cline{2-6}
               & \textbf{Self-Attn} & \textbf{End-to-end} & \textbf{FID} $\downarrow$ & \textbf{CLIP} $\uparrow$ & \textbf{IS} $\uparrow$ \\ \hline
PixArt-$\alpha$-1024 & - & - & 27.3 & 31.6 & 37.77 \\
\hline
ToMeSD (25\% ratio)~\cite{bolya2023token} & - & -7\% & 174.6 & 30.2 & 11.68 \\
DiffPruning~\cite{fang2023structural} & - & -9\% & 34.6 & 32.0 & - \\
EcoDiff~\cite{zhang2024effortless} & - & -9\% & 32.2 & 32.0 & - \\
DeepCache (N=2)~\cite{ma2024deepcache} & - & -25\% & 31.6 & 33.1 & 37.44 \\
\hline
% 13K

\methodname ($r= 40\%$)                & -48\%  & -24\%  & 28.7  & 32.1 & 37.96 \\
\methodname ($r= 40\%$) + distillation  & -48\%  & -24\%  & 27.4 & 31.9 & 36.85 \\
% 15k
\methodname ($r= 45\%$)               & -55\%  & -28\%  & 29.3 & 31.9 & 36.67\\
\methodname ($r= 45\%$) + distillation  & -55\%  & -28\%  & 28.5 & 31.9 & 36.65 \\
\hline
\end{tabular}
\vspace{-5pt}
\caption{\small Results on PixArt-$\alpha$-1024, where $r$ denotes the average pruning ratio (across blocks and timesteps) in \methodname.}\label{table:result_pixart_1024_quality}
\vspace{-1pt}
\end{table*}

\begin{table*}[t]
\centering
\small
\begin{tabular}{|c|c|c|c|c|c|c|}
\hline
\multirow{2}{*}{\textbf{Model}} & \multicolumn{2}{c|}{\textbf{FLOPs Reduction}} & \multicolumn{2}{c|}{\textbf{Latency (Efficient Attn)}} & \multicolumn{2}{c|}{\textbf{Latency (Native Attn)}} \\ \cline{2-7}
               & \textbf{Self-Attn} & \textbf{End-to-end} & \textbf{Self-Attn} & \textbf{End-to-end} & \textbf{Self-Attn} & \textbf{End-to-end} \\ \hline
PixArt-$\alpha$-1024      & -      & -      & 0.68s      & 1.68s      & 2.16s      & 3.17s \\ \hline
\methodname ($r= 45\%$)       & -55\%  & -28\%  & \makecell{0.50s \\ (-26\%)} & \makecell{1.50s \\ (-11\%)} & \makecell{1.36s \\(-37\%)} & \makecell{2.38s \\ (-25\%)} \\ \hline
\end{tabular}
\vspace{-5pt}
\caption{\small Comparison of FLOPs and latency on Nvidia H100 GPUs, measured with batch size 64.}\label{table:result_pixart_1024_gpu_latency}
% \vspace{-7pt}
\end{table*}

% \begin{table}[h!]
% \centering
% \begin{tabular}{|c|c|c|}
% \hline
% \textbf{Input Resolution} & \textbf{Model} & \textbf{Latency} \\ \hline
% \multirow{2}{*}{1024x1024} 
%   & Original & 476ms \\ \cline{2-3}
%   & Pruning 50\% tokens & 435ms (1.09x speedup) \\ \hline
% \multirow{3}{*}{1600x1600} 
%   & Original & 2481ms \\ \cline{2-3}
%   & Pruning 40\% tokens & 1619ms (1.53x speedup) \\ \cline{2-3}
%   & Pruning 50\% tokens & 1414ms (1.75x speedup) \\ \hline
% \end{tabular}
% \caption{On-device latency \hl{TODO (Dave): Replace with a Figure}}
% \label{table:result_pixart_1024_npu_latency}
% \end{table}

% \begin{table}[h!]
% \centering
% \begin{tabular}{|c|c|c|}
% \hline
% \textbf{Input Resolution} & \textbf{Baseline} & \textbf{Ours} \\ \hline
% 1024 &	476 & 440 \\
% 1152 &	670 & 575 \\
% 1280 &	823 & 732 \\
% 1408 &	1191 & 930 \\
% 1536 &	1621 & 1300 \\
% 1600 &  2436 & 1414 \\
% 1664 &	OOM & 1501 \\
% 1792 &	OOM & 1937 \\
% 1920 &	OOM & 2355 \\
% \hline
% \end{tabular}
% \caption{On-device latency \hl{TODO (Dave): Replace with a Figure}}
% \label{table:result_pixart_1024_npu_latency}
% \end{table}

\begin{figure*}
    \centering
    \begin{subfigure}[t]{\linewidth}    
        \centering
        \includegraphics[width=0.98\linewidth]{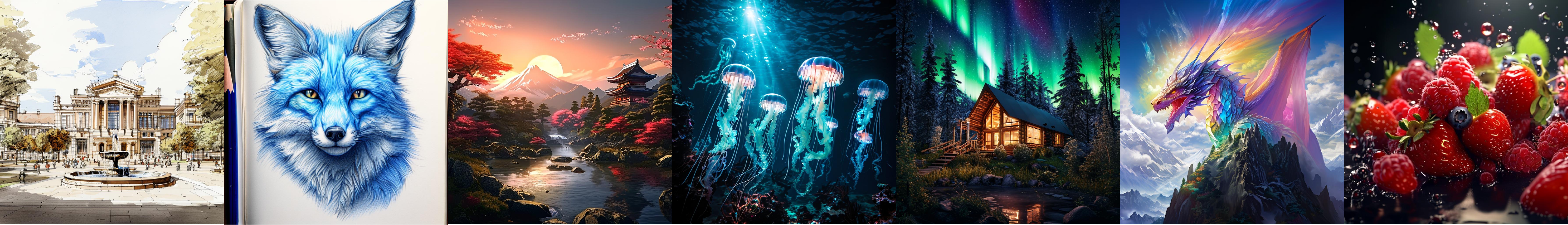}
        % \vspace{-5pt}
        \caption{\small Pretrained PixArt-$\alpha$-1024 model~\cite{chen2023pixart}.}
    \end{subfigure}
    % \vspace{1pt}
    \begin{subfigure}[t]{\linewidth}    
        \centering
        \includegraphics[width=0.98\linewidth]{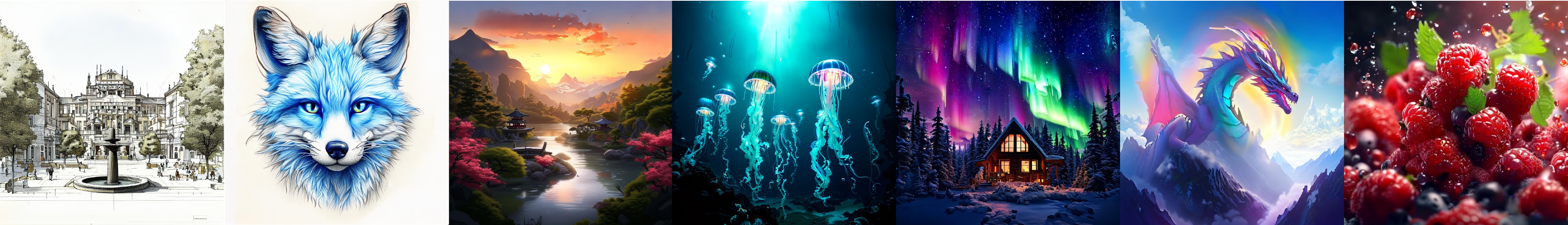}
        % \vspace{-5pt}
        \caption{\small Proposed \methodname with 40\% FLOPs Pruning. The semantics are preserved and no additional artifacts. }
    \end{subfigure}
    % \vspace{-8pt}
    \caption{\small Qualitative comparison to PixArt-$\alpha$-1024.}\label{fig:result_qualitative_comparison}
    % \vspace{-3pt}
\end{figure*}

\begin{figure}[t]
    \centering
    \includegraphics[width=0.9\linewidth]{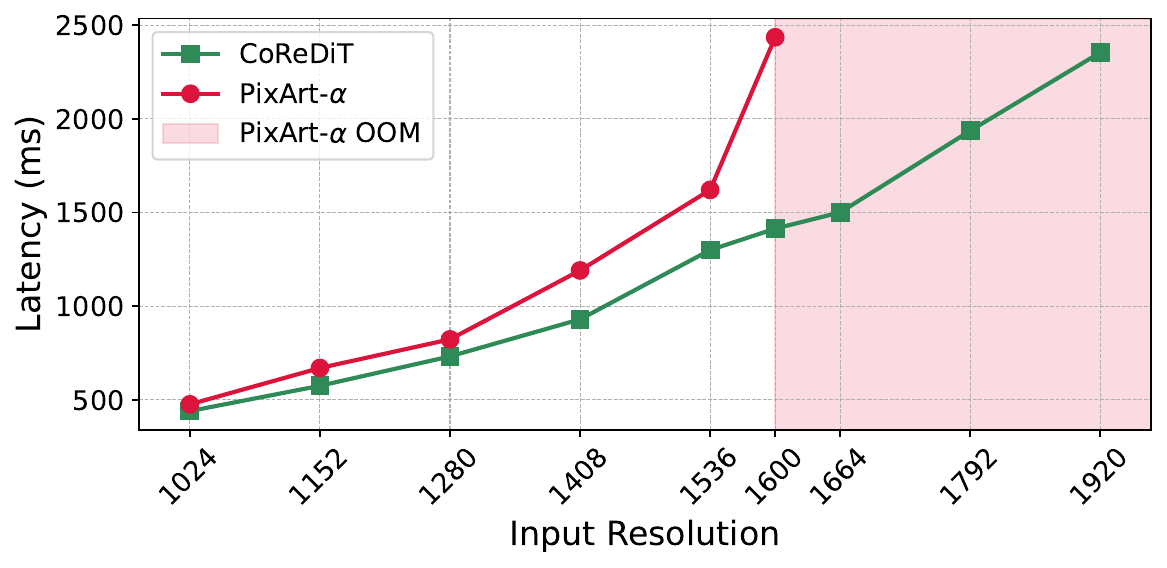}
    \vspace{-1em}
    \caption{\small Per-block latency comparison on Qualcomm Snapdragon 8 Elite NPUs. OOM: out-of-memory.}
    \label{fig:result_pixart_1024_npu_latency}
    % \vspace{0pt}
\end{figure}

\paragraph{Quality metrics.}
\cref{table:result_pixart_1024_quality} compares PixArt-$\alpha$-1024 with representative token-reduction and caching baselines~\cite{bolya2023token, fang2023structural, zhang2024effortless, ma2024deepcache}.
At $r=40\%$ average pruning, \methodname reduces self-attention FLOPs by 48\% (24\% total FLOPs) while maintaining quality close to the baseline (FID 28.7 vs. 27.3). Notably, CLIP and IS remain competitive, indicating that semantic alignment is preserved under substantial attention compute reduction.
%
% without distillation, FID rises modestly (27.3 vs 28.7), and both CLIP and IS improve.
%
With distillation, \methodname further narrows the FID gap (27.4) with only minor changes in IS.
%
% Adding distillation recovers gap of FID (27.4) with small drops in IS (36.85), demonstrating minimal perceptual or semantic loss.
%
\cref{fig:result_qualitative_comparison} shows that \methodname preserves global semantics and fine details without introducing noticeable artifacts at 40\% pruning.

\paragraph{Efficiency and latency.} 
\cref{table:result_pixart_1024_gpu_latency} reports both self-attention and end-to-end latency on Nvidia H100 GPUs under two attention backends: (i) highly-optimized kernels \textsc{xformers.memory\_efficient\_attention} (Efficient Attn) and (ii) PyTorch native attention implementations (Native Attn).
With 45\% pruning, \methodname reduces self-attention latency by 26\% under the efficient attention backend, but end-to-end latency improves by 11\% because non-attention components (e.g., FFN/MLP, I/O, and kernel launch overheads) remain unchanged.
Under native attention, latency improvements track the reduced quadratic attention cost more closely (37\% in self-attention and 25\% end-to-end).
% 
% the gains track FLOPs more closely: 37\% for self-attention and 25\% for end-to-end latency.
%
% The smaller percentage speedup reflects highly optimized kernels that do not scale quadratically with token count, while the native path exposes more of the quadratic attention savings, resulting in larger relative improvements.

% separating as a new paragraph as we want people to not miss this result :)
Beyond GPU inference, \cref{fig:result_pixart_1024_npu_latency} shows per-block latency on Qualcomm Snapdragon 8 Elite NPUs.
At 50\% pruning, \methodname achieves up to $1.72\times$ per-block speedup at 1600 resolution.
Notably, the baseline PixArt-$\alpha$ encounters out-of-memory (OOM) error at resolutions above 1600, whereas \methodname runs up to 1920 resolution and matches the baseline latency at 1600.
% , with compatible latency to that of PixArt-$\alpha$ at 1600 resolution.
%
% These results indicate that substantial compute and runtime savings can be achieved with minimal perceptual and semantic degradation, showing favorable efficiency–quality trade-offs.
%
These results indicate that structured attention skipping can provide not only runtime gains but also meaningful \emph{memory headroom} for higher-resolution deployment.

% pixart sigma 
% \begin{table}[t!]
% \centering
% % \small
% \begin{tabular}{|c|c|c|c|c|c|c|c|c|c|c}
% \hline
% \multirow{2}{*}{\textbf{Model}} & \multicolumn{2}{c|}{\textbf{FLOPs Reduction}} & \multicolumn{2}{c|}{\textbf{Latency}} & \multicolumn{4}{c|}{\textbf{Image Quality}} \\ \cline{2-9}
% & \textbf{Self-Attn} & \textbf{Total} & \textbf{Self-Attn} & \textbf{Total} & \textbf{FID} $\downarrow$ & \textbf{LPIPS} $\downarrow$ & \textbf{CLIP} $\uparrow$ & \textbf{IS} $\uparrow$ \\ \hline

% PixArt-$\Sigma$-2048 & - & - & 4.62s & 6.68s & 26.0 & 0.838  & 31.4  & 37.49\\ \hline
% \makecell{\methodname (23\%) \\ w/ distill}  & -32\% & -25\% & \makecell{3.61s\\ (-22\%)} & \makecell{5.67s \\ (-15\%)}  & 28.0 & 0.845 & 31.4 & 36.52 \\ \hline
% % note: pruning 23 % tokens (w/ distill)
% \end{tabular}
% \vspace{-7pt}
% %  (batch size = 32)
% \caption{\small Results on PixArt-$\Sigma$-2048, with latency measured on Nvidia H100 GPUs using Eff. Attention.}\label{fig:metrics_pixart_sigma}
% \vspace{-6pt}
% \end{table}

\begin{table*}[t!]
\centering
% \small
\begin{tabular}{|c|c|c|c|c|c|c|c|c|c|c}
\hline
\multirow{2}{*}{\textbf{Model (ratio)}} & \multicolumn{2}{c|}{\textbf{FLOPs Reduction}} & \multicolumn{2}{c|}{\textbf{Latency}} & \multicolumn{3}{c|}{\textbf{Image Quality}} \\ \cline{2-8}
& \textbf{Self-Attn} & \textbf{End-to-end} & \textbf{Self-Attn} & \textbf{End-to-end} & \textbf{FID} $\downarrow$ & \textbf{CLIP} $\uparrow$ & \textbf{IS} $\uparrow$ \\ \hline

PixArt-$\Sigma$-2048 & - & - & 4.62s & 6.68s & 26.0 & 31.4  & 37.49\\ \hline
\makecell{\methodname ($r= 23\%$) \\ + distillation}  & -32\% & -25\% & \makecell{3.61s\\ (-22\%)} & \makecell{5.67s \\ (-15\%)}  & 28.0 & 31.4 & 36.52 \\ \hline
% note: pruning 23 % tokens (w/ distill)
\end{tabular}
% \vspace{-7pt}
%  (batch size = 32)
\caption{\small Results on PixArt-$\Sigma$-2048, with latency measured on Nvidia H100 GPUs using Efficient Attention.}\label{fig:metrics_pixart_sigma}
% \vspace{-6pt}
\end{table*}

% \vspace{-1pt}
\subsection{High Resolution: PixArt-$\Sigma$-2048}
% \vspace{-5pt}
We apply \methodname to PixArt-$\Sigma$-2048 to evaluate high-resolution image generation.
As shown in \cref{fig:metrics_pixart_sigma}, at an average pruning level of 23\%, \methodname reduces self-attention FLOPs by 32\% (25\% total FLOPs), translating into a 22\% reduction in self-attention latency and a 15\% end-to-end latency reduction (batch size 32, memory-efficient attention).
Quality remains high: CLIP remains the same, while FID and IS exhibit only a modest drop  (26.0 vs. 28.0 and 37.49 vs. 36.52, respectively).
We attribute part of this gap to a resolution-domain mismatch: our fine-tuning dataset is at 1K (upscaled to 2K) while evaluation uses native 2K images.
Upscaled 1K data can under-represent 2K high-frequency details and long-range structures;
fine-tuning on native 2K images is expected to narrow the remaining quality gap.

% Here we have limitations of lack of high-resolution data.

% \begin{table}[h!]
% \centering
% \begin{tabular}{|l|c|c|c|c|c|c|c|c|}
% \hline
% \textbf{Model} & \textbf{Token Red.} & \textbf{Attn Red.} & \textbf{Fid (2048)} & \textbf{Fid (299) $\downarrow$} & \textbf{LPIPS $\downarrow$} & \textbf{CLIP $\uparrow$} & \textbf{IS $\uparrow$} \\
% \hline
% PixArt-$\Sigma$-2048  & -    & -    & 8.05  & 26.02 & 0.838  & 0.314  & 37.49 \\
% 7k steps         & 40\% & 58\% & 10.17 & 30.77 & 0.847  & 0.315  & 33.26 \\
% 6.5k steps       & 37\% & 55\% & 9.74  & 30.08 & 0.844  & 0.314  & 34.98 \\
% 6k steps         & 34\% & 50\% & 10.12 & 32.94 & 0.831  & 0.314  & 36.62 \\
% 5.5k steps       & 31\% & 46\% & 9.80  & 31.77 & 0.823  & 0.317  & 36.73 \\
% 5k steps         & 28\% & 42\% & 9.51  & 32.63 & 0.820  & 0.316  & 36.87 \\
% 4.5k steps       & 26\% & 38\% & 9.08  & 30.91 & 0.821  & 0.317  & 35.27 \\
% 4k steps w/ distill & 23\% & 34\% & 8.56 & 28.00 & 0.845 & 0.314 & 36.52 \\
% 3.5k steps w/ distill & 20\% & 31\% & 8.50 & 28.97 & 0.847 & 0.314 & 37.03 \\
% \hline
% \end{tabular}
% \caption{Performance metrics for PixArt-$\Sigma$-2048 across different pruning configurations.}\label{fig:metrics_pixart_sigma}
% \end{table}

% \textbf{Model} & \multicolumn{2}{c|}{\textbf{FLOPs}} & \multicolumn{2}{c|}{\textbf{Against Real}} & \textbf{Text-Image} \\ \cline{2-6}
%                & Self-Attn & Total & FID $\downarrow$ & LPIPS $\downarrow$ & CLIP $\uparrow$

\subsection{Video Generation: MagicDrive-V2}
% \vspace{-5pt}
% \begin{table}[t]
% \centering
% \begin{tabular}{|c|c|c|c|c|c|c|}
% \hline
% \textbf{Model} & \textbf{FVD $\downarrow$} & \textbf{LPIPS $\downarrow$} & \textbf{PSNR $\uparrow$} & \textbf{SSIM $\uparrow$} & \textbf{mAP $\uparrow$} & \textbf{mIoU $\uparrow$} \\
% \hline
% MagicDrive-V2 & - & - & - & - & - & - \\
% \methodname (TODO\%) & - & - & - & - & - & - \\
% \hline
% \end{tabular}
% \caption{Results on MagicDrive-V2.}\hl{TODO (Dave): Update table here}
% \label{table:result_magicdrive}
% \end{table}

\begin{table*}[t!]
\centering
\small 
\begin{tabular}{|c|c|c|c|c|c|c|c|c|}
\hline
\multirow{2}{*}{\textbf{Model}} & \multicolumn{2}{c|}{\textbf{FLOPs Reduction}} & \multicolumn{4}{c|}{\textbf{Video Quality}} & \multicolumn{2}{c|}{\textbf{Cond. Alignment}} \\
\cline{2-9}
& \textbf{Self-Attn} & \textbf{End-to-end} & \textbf{PSNR $\uparrow$} & \textbf{LPIPS $\downarrow$} & \textbf{SSIM $\uparrow$} & \textbf{FVD $\downarrow$} & \textbf{mAP $\uparrow$} & \textbf{mIoU $\uparrow$} \\
\hline
MagicDrive-V2 & - & - & 14.28  & 0.422 & 0.372 & 107.8 & 18.4\% & 21.5\% \\ \hline
% t20 (10K)
% \methodname (17\%) & - & 14.24 & 0.424 & 0.374 & 111.4 & 18.5\% & 21.2\% \\
% t20 (15k)
\methodname ($r= 26\%$) & -39\% & -8\% & 14.25 & 0.424 & 0.378 & 119.8 & 18.1\% & 21.2\% \\
% t15 (10k)
% \methodname (23\%) & \hl{-7\%} & 14.27 & 0.425 & 0.378 & 120.7 & 18.2\% & 21.1\% \\
% t15 (15k)
% \methodname (34\%) & - & 14.25 & 0.428 & 0.381 & 136.5 & 17.3\% & 20.8\% \\
\hline
\end{tabular}
% \vspace{-7pt}
\caption{\small Video generation results on video quality and conditional alignment via 3D object detection.}\label{table:result_magicdrive}
\end{table*}

% As shown in \cref{table:result_magicdrive}, on MagicDrive-V2, \methodname preserves frame-level fidelity.
% while slightly weakening temporal coherence.
%
% As current research in generation is trending in higher resolution and longer sequences. The use of DiT architecture is inevitable. The video generation architecture is relying on higher token length and exponential growth in computation power. Therefore, it is an urgent to reduce unnecessary computation. 

% Recent trends in video generation research emphasize higher resolutions and longer sequences, which lead to increased token sequences and substantial growth in computational demands.
%
We extend \methodname, which is broadly applicable across vision tasks and agnostic to input modalities, to the state-of-the-art video generation DiT for autonomous driving - MagicDrive-V2~\cite{gao2024magicdrive}.
We prune only spatial transformer blocks and achieve a 39\% self-attention FLOPs reduction (8\% total FLOPs at the full-model level) as shown in~\cref{table:result_magicdrive}.
%
% Our integration demonstrates that redundancy in the model can be reduced. Specifically, we achieve a 39\% FLOPs reduction in self-attention within the spatial transformer blocks, while preserving visual quality. 
%
Despite this reduction, \methodname preserves strong per-frame quality (PSNR, LPIPS, and SSIM remain clost to the baseline) and maintains conditional aligment, as measured by BEVFormer mAP and mIoU.
%
% Additionally, mAP and mIoU also suggest that our method provides stable conditional alignment. 

Currently, \methodname is applied only to spatial transformer blocks in which each frame is processed independently. Consequently, we observed a drop in FVD, which reflects temporal consistency in the generated video.
Future work will focus on integrating temporal consistency objectives, such as cross-frame coherence, to mitigate this limitation. We expect such enhancements will improve FVD performance while maintaining the computational efficiency demonstrated by our current approach.
\vspace{-5pt}
% Potential improvement (moved to appendix):
%
% One reason is that our pruning method is applied only to the spatial transformer blocks, with each frame processed independently, potentially leading to discrepancies between frames.
%
%
% Optional: defense for FVD drop
% In addition, we note a limitation of FVD evaluation: we observe a substantial distribution shift from Kinetics-400 which is a human action recognition dataset (used to train the I3D~\cite{carreira2017quo} feature extractor) to nuScenes which is a driving dataset (used to fine-tuning MagicDrive-V2).
%
% This mismatch leads to suboptimal feature representations when using I3D, causing the FVD metric to mischaracterize the perceptual and semantic quality of generated driving videos.
%
% In particular, we find that the FVD changes drastically during fine-tuning, suggesting that the metric is highly sensitive in representation space that may not correspond to meaningful improvements in visual quality.

\subsection{Ablation Study}
% \vspace{-5pt}

% \begin{table}[t]
% \centering
% \small 
% \begin{tabular}{|c|c|c|c|c|c|}
% \hline
% \textbf{Experiment} &\multicolumn{4}{c|}{\textbf{Image Quality}} \\
% \cline{2-5}
%  & \textbf{FID $\downarrow$} & \textbf{LPIPS $\downarrow$} & \textbf{CLIP $\uparrow$} & \textbf{IS $\uparrow$} \\
% \hline
% PixArt-$\alpha$-1024 & 27.3 & 0.810 & 31.6 & 37.77 \\
% \hline
% % Prune 15k steps & 45\% & 29.3 & 0.812 & 0.319 & 36.67 \\
% \methodname (45\%) & 28.5 & 0.815 & 31.9 & 36.65 \\
% % Prune 14k steps & 42\% & 28.5 & 0.811 & 0.320 & 38.29 \\
% % \methodname (-42\%) w/ \herbert{should be "without" here?} distill & 28.3 & 0.805 & 0.319 & 37.42 \\
% \hline
% % 7.5k
% Random selection (23\%) & 644.5 & 0.979 & 21.5 & 1.0 \\
% % 8k
% Disable reconstruction (30\%) & 30.4 & 0.808 & 31.7 & 34.20 \\
% %
% Uniform ratio (41\%) & 32.3 & 0.812 & 31.5 & 33.68 \\
% % Uniform ratio (41\%) w/o distill & 30.6 & TODO & 31.8 & 35.48 \\
% \hline
% \end{tabular}
% \vspace{-7pt}
% \caption{\small Ablation study on PixArt-$\alpha$-1024, all with distillation.}\label{table:ablation_study}
% \vspace{-10pt}
% \end{table}

\begin{table}[t]
\centering
\small 
\begin{tabular}{|c|c|c|c|}
\hline
\multirow{2}{*}{\textbf{Experiment}} &\multicolumn{3}{c|}{\textbf{Image Quality}} \\
\cline{2-4}
 & \textbf{FID $\downarrow$} & \textbf{CLIP $\uparrow$} & \textbf{IS $\uparrow$} \\
\hline
PixArt-$\alpha$-1024 & 27.3 & 31.6 & 37.77 \\
\hline
\methodname ($r= 45\%$) & 28.5 & 31.9 & 36.65 \\
\hline
% 7.5k
Random selection ($r= 23\%$) & 644.5 & 21.5 & 1.00 \\
% 8k
Disable reconstruction ($r= 30\%$) & 30.4 & 31.7 & 34.20 \\
Uniform ratio ($r= 41\%$) & 32.3 & 31.5 & 33.68 \\
\hline
\end{tabular}
% \vspace{-7pt}
\caption{\small Ablation study on PixArt-$\alpha$-1024, with distillation in all experiments.}\label{table:ablation_study}
% \vspace{-10pt}
\end{table}

\cref{table:ablation_study} presents our ablation studies to evaluate key components of \methodname on PixArt-$\alpha$-1024 (all with distillation):
(1) \textit{Random Token Selection.} To evaluate the importance of spatial-coherence-based token selection, we conduct experiments of randomly sampling tokens to skip in each block; this yields extremely poor quality (FID 644.5 at 23\% pruning), proving that spatial-coherence based selection is effective.
(2) \textit{Skipping Tokens without Reconstruction.} To illustrate the contribution of  reconstruction, we disable it and simply skip selected tokens. At 30\% pruning ratio, FID degrades to 30.4, indicating that reconstruction is necessary to preserve semantic fidelity.
(3) \textit{Uniform Pruning then Fine-tuning.} To justify the effectiveness of progressive pruning, we apply a uniform 50\% per-block pruning to the pretrained model and then fine-tune; this results in worse FID 32.3 at average $40.6\%$ pruning, highlighting the advantage of progressive, block-adaptive pruning.

\section{Conclusion}
% \vspace{-5pt}

We introduce \methodname, a token pruning framework for DiTs that exploits spatial coherence as a practical signal of local redundancy.
\methodname combines (1) an efficient pre-attention selector that skips high-coherence tokens, (2) a coherence-guided reconstruction operator that preserves a dense token lattice and mitigates pruning artifacts, and (3) a progressive, block-adaptive pruning schedule that allocates pruning where redundancy is highest across blocks and denoising steps.
Experiments on state-of-the-art image and video generators demonstrate substantial reductions in self-attention compute and measurable end-to-end speedups on both cloud GPUs and mobile NPUs, while maintaining strong perceptual quality and conditional alignment.
In future work, we plan to (1) further reduce non-attention overhead by extending efficiency to FFN/MLP components, and (2) improve video temporal consistency by incorporating cross-frame coherence objectives and pruning strategies that explicitly model temporal redundancy.


{
    \small
    \bibliographystyle{ieeenat_fullname}
    \bibliography{main}
}

% \input{sections/appendix}

\end{document}